\documentclass{article}
\usepackage{graphicx,amsmath,longtable}

\begin{document}

\title{Visualization of Jacques Lacan's Registers of the Psychoanalytic 
Field, and Discovery of Metaphor and of Metonymy. Analytical Case Study 
of Edgar Allan Poe's ``The Purloined Letter''}

\author{Fionn Murtagh, Giuseppe Iurato \\
Email: fmurtagh@acm.org}

\maketitle

\begin{abstract}
We start with a description of Lacan's work that we then take into our 
analytics methodology. In a first investigation, a Lacan-motivated template 
of the Poe story is fitted to the data. A segmentation of the storyline is 
used in order to map out the diachrony. Based on this, it will be shown how 
synchronous aspects, potentially related to Lacanian registers, can be sought. 
This demonstrates the effectiveness of an approach based on a model template of
the storyline narrative. In a second and more comprehensive investigation, we 
develop an approach for revealing, that is, uncovering, Lacanian register 
relationships. Objectives of this work include the wide and general application 
of our methodology. This methodology is strongly based on the ``letting the data 
speak'' Correspondence Analysis analytics platform of Jean-Paul Benz\'ecri, that is 
also the geometric data analysis, both qualitative and quantitative analytics, 
developed by Pierre Bourdieu.
\end{abstract}


\section{General Objectives and Outline}

Narrative analysis is at issue here, using what has been a highly profiled text 
in literary studies. Our approach is unsupervised, relative to supervised learning. 
It involves visualization of our data, using semantic content, in such a way that 
there is revealing of relationships in the data. This can be taken further, if so 
desired, in the direction of statistical modelling and supervised machine learning. 
Our desire though is to deal with dynamic and fluid expression, and flow and 
evolution in our data content. Therefore our analysis methodology is motivated very 
much by the visualization and the verbalization of our data (Blasius and Greenacre, 
2014). We can even state (see subsection 2.5) that our methodology is data analysis 
integrated with information synthesis. 

A prime objective in this work is the introducing of innovative
potential, as both language studies and psychoanalysis still tend to
reject quantifying approaches to text and psyche respectively.
However, this is more due to historical differentiation within the
sciences than to ultimate scientific justification. Lacan's further
development of psychoanalysis and the symbolic is affected by 
L\'evi-Strauss' structuralist reasoning, thus necessitating systematic
approaches of visualizing these very structures. Structuralism and
geometric data analysis share a common epistemology as both are
genuinely relational or topological in their ways of conceptualizing
entities. As a matter of fact, the first-ever presentation of
correspondence analysis by Jean-Paul Benz\'ecri dealt with textual data
and the visualization of the structures that are hidden in texts. Thus, the
main orientation of this paper is both innovative in a methodological
sense and orthodox in an epistemological sense. In doing so, and due to
the traceability of the methodic steps, it might invite scholars from
language studies and psychoanalysis to apply similar strategies to their
own research fields. (Cf.\ Acknowledgements.)

Our methodology is also highly cross-disciplinary. We aim to demonstrate how an 
important set of perspectives, developed by Lacan, are generally and broadly 
applicable, to literary theory. But an even greater objective is to apply Lacan's 
work to practical investigative problem-solving, including psychoanalytical 
investigative work. Such is our ultimate aim and ambition. 

In section 2, comprehensive background discussion is provided on Lacan's revealing 
and relevant methodology. 

An initial study that is exploratory is carried out in section 3. Geometric data 
analysis is our methodology, based on the work of eminent social scientist, Pierre 
Bourdieu, who followed in his work, eminent data scientist, Jean-Paul Benz\'ecri, 
whose earliest work embraced mathematics and linguistics. 

In section 4, we show how Lacanian registers can be visualized in the context of 
narrative flow. The culmination of such visualization is in subsection 4.5.

Then in section 5 we seek analytical perspectives that will be revealing in regard 
to discovering metaphor and metonymy. Synonyms in a semantic framework are of 
potential relevance. Word associations are analysed through clusters determined 
from the semantic mapping. 

In the Conclusions, section 6, a link is provided to the Poe story following our 
preprocessing of it with each successive sentence on a new line. Furthermore a 
good part of the R software used in this work is provided. 

\section{Introduction to Lacan's Registers and Analytical Framework}

\subsection{Jacques Lacan's Language and Unconscious: Observing and 
Tracking the Imaginary and the Symbolic, Dynamically Engaged with the Real}

Lacan's reading of Poe's story is closely based (although not explicitly
indicated by Lacan) on Marie Bonaparte's reading of the same Poe's
story, according to which the letter Dupin finds hanging in a letterholder
between the ``cheeks'' of the fireplace represents the
``rephallization'' of the mother (i.e., the queen, with the letter as the
phallus lacking to the queen). This has been just the main remark due
to Derrida in the 1980s (Derrida 1980). See also Todd (1990, Part V,
Ch. VIII, pp. 166-171).

Jacques Lacan's seminar on this story by Edgar Allan Poe (Lacan, 1956) includes 
interpretation that is ``sufficient for us to discern [...] so perfect a 
verisimilitude that it may be said that truth here reveals its fictive arrangement''. 
A dialogue in the tale by Poe ``presents the real complexity of what is ordinarily 
simplified, with the most confused results, in the notion of communication''. 
Complexity results from: ``... communication is not transmissible in symbolic 
form. It may be maintained only in the relation with the object.'' What is integral 
to this: ``Language delivers its judgement to whoever knows how to hear it''.
 
Then: ``What Freud teaches us in the text we are commenting on is that the subject 
must pass through the channels of the symbolic, but what is illustrated here is 
more gripping still: it is not only the subject, but the subjects, grasped in their 
intersubjectivity, [...] who [...] model their very being on the moment of the 
signifying chain which traverses them.''

The context, that we are dealing with, is simple if we just look at events from 
afar. Consider the following, in Lacan's seminar. The signifier related to this 
purloined letter: ``It remains for it now only to answer that very question, of 
what remains of a signifier when it has no more signification.'' This is our 
interest too, even if: ``what the `purloined letter' nay, the `letter in sufferance,' 
means is that a letter always arrives at its destination''
.
We now turn to the following work, Ragland (2015), in order to point out motivation 
and justification for what follows in this article, i.e., the general and potentially 
very beneficial and rewarding application. 

Ragland (2015) presents a comprehensive account of Lacan's work. Let us summarize 
the important perspectives for us in this work. All citations in the following part 
of this subsection are from Ragland (2015). At issue is how Lacan provides a 
``conceptualization of mind structure'' (p.\ 107), and that (p.\ 112) ``Lacan gives 
us a means to go beyond biological or cultural materialisms.'' Therefore (p.\ 137) 
``Lacan was concerned with structure [...] -- not with the {\em content} of the 
unconscious.''
 
Language is both representation and also an instrument: ``Language itself merely 
represents desire at one remove'' (p.\ 3). ``language promotes jouissance, not just 
communication, or information'' (p.\ 2). Thus (p.\ 62), ``speech (or writing) carry 
desire [...] `discourse' is not grammar''. In this perspective, then (p.\ 5), ``for 
Lacan [...] language is imposed from the outside. It is not innate or hardwired into 
the brain.''

Fundamental to Lacan is that language's way of being an infrastructure for desire, 
is manifested visually and by shape, as summarized in the following terms (p.\ 51). 
``We have left behind the linear logic of linguistics and phenomenology and walked 
into the universe of multiform, contradictory logic that Lacan calls a way of 
`topologizing'.'' For Lacan, one's mind is related to one's body, and the topology 
of the body are in particular what comprise holes, input and output, and the body 
surface (cf.\ p.\ 21). Important then for the mind is when body parts, and objects 
later, are perceived as missing, or are gaps. We have (p.\ 106): ``the surface of the
body marked by holes and rims (mouth, ear, nose, etc.)''

``For psychoanalysis, topology [...] is not a metaphor, but confirms the presence of 
the real ... Topology is an active showing of the real of structure.'' (p.\ 120). 
``Topology [...] is not a metaphor. Not an allegory. It does not represent the 
subject. [...] Topology presents `the foundations of the subject's position.' The 
subject combines itself in the Borromean unit” (pp. 124--125). There follows how 
important the subject's gaze, and ``visual structure'' are here. 

For Lacan, ``language is duplicitous'', and it expresses ``affective knowledge'', and 
justification for this is how language is not a set, fixed (grammatical or taxonomic) 
structure, but is dynamical, and fluid (p.\ 114): ``language is duplicitous, not only 
because it is an agent of {\em repression}, but also because it does not succeed in 
repressing the material of identifications that aim the drives towards lures, 
towards the goal of repeating the familiar.'' ``By bringing the drives into 
language as an affective knowledge -- a montage of the real and the imaginary, 
the symbolic and imaginary, the real and symbolic -- conflict or {\em torsion} can be 
proposed as a property of language whose referent is the concrete nature of the 
drives.'' (pp. 115--116).

``In consideration of the predilection among intellectuals to think of language in 
purely abstract logical terms, it is vital that we become aware that there is a 
dynamics of language.'' (pp.\ 116--117). Ultimately, at issue (p.\ 127) is how we 
have ``language as signifying something other than what it says.''

Regarding ``geometric'' written in the following manner, it is stated that: ``thought 
and body are geo-metric'' (p.\ 12), ``topological structure is a knowledge of being, 
not an academic knowledge” (p.\ 4), and (p.\ 108) ``topology is not [...] a knowledge 
to be taught by concepts or fundamental texts: `It is a practice of the hole and its 
edge' ''

``As a mathematical knowledge of the real, topology itself draws `pictures' of how 
body, language, and world co-exist, intertwined in contradictory ways, that can be 
explained logically all the same.'' (p.\ 45). At issue is ``Lacan's theory of a 
topological ‘structuring’ of the unconscious'' (p.\ 46). ``Lacan's topological 
forms [...] introduces the real into language, as a set of affective, albeit 
emotionally ungraspable, meanings.'' (p.\ 112).

Metaphor and metonymy are quite central issues here: ``each of Lacan's discourse 
structures has [...] the double, substitutive structure of metaphor'' (p.\ 57), 
and ``interpretation works as a metaphor which allows substitution of one thing
for another''(p.\ 5), with this perspective (p.\ 88): ``the dialectical link 
between metaphor and metonymy as poetic tropes that make the brain function'', 
``He described condensation as metaphors (substitutes) linked to the object that 
causes desire [...] by the concrete metonymous displacements and contiguities of 
desire.''

For Lacan, ``structure is Borromean. By Borromean he means the knotting together 
of the symbolic, imaginary, and real dimensions bound together by the 
symptom/{\em sinthome}.'' (p.\ 24). There are: ``the three different dimensions of 
knowledge -- the real, the symbolic, the imaginary'' (p.\ 2).

So there is the following, with a major role in our work that will follow: 
``the signifying chain is not [...] grammar, language, or writing, but, rather, a 
chain of dimensions --- Real, Symbolic, Imaginary'' (p.\ 63). Thus ``the real of 
the (partial) drives [...] what is repressed in the real returns, anyway, into 
the symbolic order of language and social conventions.'' (p.\ 106). ``The drive 
[...] designates the prevalence of an `organic' dimension of symbolic and imaginary 
traits that coalesce with the real of the flesh as a mapping in language'' (p.\ 106).

While fully separate from what will follow in our analyses of Poe's ``The purloined 
letter'', it is nice to note, in the paragraph to follow, how Lacan also used the term 
``letter''! 
 
``Ancient cave paintings were a writing before writing, a way to make a meaningness 
Lacan called the letter.'' (p.\ 12). This expresses a role for this term, letter: 
``every truth has the structure of the fiction and that truth and fiction are linked 
by the {\em letter}. The function of the letter is topological. ... it indicates the {\em place}
where language and the unconscious are linked.'' (p.\ 12). Here, there is: ``a `letter' 
or visual representation'' (p.\ 127). ``Lacan described the {\em lettre} as a {\em place} where 
being ({\em l'\^etre}) resides between the unconscious and language, calling the {\em lettre} a 
localized signifier that one can recognize as language converging with the 
unconscious.'' (p.\ 136).

\subsection{Psychoanalytical Use of Edgar Allan Poe Story}

Edgar Allan Poe's ``The Purloined Letter'' (Poe, 1845a, 1845b, 1845c) is a story that 
is investigative and elaborative. It is not just explanatory, reducing the case study 
in this story to facts and assertions that are ordered. Rather, it is also elucidatory, 
and positioning in a larger, broader, contextual picture. Also this allows to identify 
better where truth lies by means of simple psychoanalytic tools. Positioning is done 
through contextualization. We hypothesize that any and all such elucidation, and 
contextual positioning, is potentially relevant for various domains such as theatre and 
drama, legends and mythology, and {\em mutatis mutandis}, for poetry and music.

Therefore, Poe's story is not simply the investigation of illegal behaviour. There are 
parallels and analogies drawn with schoolboys playing with marbles and strangely enough 
with mathematical reasoning. These strange connections are just possible in the 
unconscious realm. A lot of foremost thinkers have discovered, or at least viewed, 
very interesting mappings of Poe's story into the most interesting contexts. See 
Department of English at FJU (2010) for discussion with graphical portrayal of Michel 
Foucault, Jacques Lacan, Jacques Derrida and others.

Description follows of the psychoanalytical approach developed by Lacan, encompassing 
analysis of synchrony and of diachrony. Diachrony can be based on the inducing of a 
segmentation of the narrative or storyline into a sequence of main scenes or acts. 
The synchronous elements decompose any act by means of the three Lacanian registers 
or orders of the so-called {\em psychoanalytic field} in which every human event performs 
at the unconscious level. The three Lacanian registers, comprising the psychoanalytic 
field, are the real, the imaginary and the symbolic. Lacanian psychoanalysis seeks to 
outline the co-participation of these three registers in each event and subject of the 
story, but with a synchronic predominance of one over the others, which will then be 
the one that is diachronically identifiable.  However this is only under the surface of
symbolic, the only register that represents the other remaining two
(Recalcati 2012--16, Vol. II, pp. 549--550).

Our study has the following objective. Firstly we seek to reveal or to determine 
Lacan's registers in a highly realistic case study.

Our mapping of Lacan's registers in the Poe story leads to visualization, to 
represent visually these registers, in the context of their roles. Specifically 
seeking metaphor and metonymy is at issue in a later section.

\subsection{Source of Data and Preparation}

In this subsection, and throughout this paper, we detail the data processing carried 
out, firstly for reproducibility of this study, and secondly for all aspects relating 
to generalization of this work, and application to other textually expressed content. 

The Edgar Allan Poe text of ``The Purloined Letter'' was taken from Poe (1845a). 
Accented characters required correction, following the 1845 editions in Poe (1845a, 
1845b, 1845c). A program was run on this text that determined sentence boundary 
(using a full stop), and also took into account blank lines that indicated paragraph 
boundary. Some cases of repeated dashes, repeated dots, exclamation marks and question 
marks were modified manually in the input text. The processing allows the specification 
of standard contractions that are not to be taken as sentence boundaries. (The 
following were at issue in regard to being ended with a full stop or period but this 
did not connote the end of a sentence: {\tt no}, {\tt No}, {\tt C}, {\tt G}, {\tt St}.) 
A CSV (comma separated 
values) formatted file was created, with the sentence sequence number, the paragraph 
sequence number, and the sentence content. This led to 321 sentences and 123 paragraphs. 
For each paragraph, the speaker was also noted: the Narrator, Dupin and the Prefect. 
In section 2.4, some further background description on the Poe story will be provided. 

\subsection{Dramatis Personae}

The characters in this short story are as follows: (1) C. Auguste Dupin (young 
private detective); (2) Monsieur ``G -- --'', or G.\ or Prefect (police chief); (3) 
the narrator (Dupin's friend and roommate); (4) the Minister ``D -- --'', or 
``the D -- --'', or the minister (the villain); (5 and 6) the personage [in the 
royal boudoir], or other unnamed royal person (often considered as Queen, King); 
and (7) ``S --'', sender of the letter (only one occurrence of this name).

Examples follow of the first and the last sentences. 

\begin{itemize}
\item First: ``At Paris, just after dark one gusty evening in the autumn of 18 -- , 
I was enjoying the twofold luxury of meditation and a meerschaum, in company with my 
friend, C.\ Auguste Dupin, in his little back library, or book-closet, au troisi\`eme, 
No.\ 33 Rue Dun\^ot, Faubourg St. Germain.''
\item Last: ``They are to be found in Cr\'ebillon's 'Atr\'ee''
\end{itemize} 

\subsection{Brief Background on the Geometric Data Analysis Methodology}

Our approach is influenced by how the leading social scientist, Pierre Bourdieu, 
used the most effective inductive analytics developed by Jean-Paul Benz\'ecri. See 
Le Roux and Rouanet (2004), Grenfell and Lebaron (2014), Lebaron and Roux (2015). 
This family of geometric data analysis methodologies, centrally based on 
Correspondence Analysis encompassing hierarchical clustering, and statistical 
modelling, not only organises the analysis methodology and domain of application, 
but even integrates them. The second in a set of principles for data analytics, 
listed in Benz\'ecri (1973, page 6), included the following: ``The model should 
follow the data, and not the reverse. ... What we need is a rigorous method that 
extracts structures from data.'' Closely coupled to this is that (Benz\'ecri, 1983) 
``data synthesis'' could be considered as equally if not more important relative to 
``data analysis''. Analysis and synthesis of data and information obviously go hand 
in hand.

The work of Andreas Schmitz, dealing with Angst and fear (Schmitz, 2015, Schmitz and 
Bayer, 2014), links together Freud and Bourdieu, for example, in regard to ``libido 
within habitus-field theory''. Among the conclusions in Schmitz (2015) are how we have: 

\begin{enumerate}
\item ``Libido constitutive for the foundational concepts of {\em habitus} and {\em fields}''. 
\item Janus-faced character of libido: interest and Angst as constitutive moments of 
(i) Habitus and practice, (ii) Social space and social fields, and (iii) Symbolic 
domination. 
\end{enumerate}

In Schmitz and Bayer (2014), also presented in Schmitz (2015), the limits are noted 
of statistical linear modelling for relating personality factors in social space. 
Moving beyond that methodology, there is categorical interest and personality types, 
accompanying the socio-structural information for the geometric construction of social 
space. The aim is to demonstrate in general, whether psychological characteristics 
will correspond with the structure of social space in a discontinuous way. (This 
summarizes perspectives in Schmitz and Bayer, 2014, p.\ 11. The following is from 
p.\ 14.) Habitus defines the nexus between structure and subject, whereby the 
correspondence of social position and ``psychic'' disposition are understood as 
class-specific, and thus discontinuous. Psychiatric indicators are used in a 
discontinuous way (as befits such categorical variables). From a psychoanalytic 
viewpoint, the habitus roughly corresponds to Freudian super-ego agency, hence it 
belongs to the Lacanian symbolic register. So, studying the latter, we might infer 
features of habitus, hence answer to the above issue regarding links between 
psychological characteristics and social structure. We shall focus on the linguistic.

\section{First Exploratory Study: Analysis Using Simple Diachronic Model}

Below, in this paper, most of the set of words in the Poe text are used. This is so 
as to take account of emotion and sentiment, expressed language-wise through 
adjectives and adverbs, and so on. Also below, text-based, i.e. data-based, story 
or narrative flows are considered. 

In this first study, a somewhat simplified diachronic model of the Poe story is 
used. That is, a model of the evolution or flow of the story is used. This is 
strongly based on a Lacanian interpretation. Also in this first study, from the 
text of the Poe story, nouns are used. This is in order to have a relatively quick, 
first view of the relationship between key terms. 

We consider now, the Lacanian motivation, and indeed justification, for this work.

Lacan's psychoanalytic {\em field} is structured into three dimensions or orders, termed 
the Lacan {\em registers}, which may be considered as components of this field, closely 
linked to each other (Borromean knot). These are the {\em symbolic register}, the 
{\em imaginary register} and the {\em real order}. The Lacan psychoanalytic field 
relies on the unconscious realm. 

The {\em symbolic register} is that field component in which {\em signifiers} act, 
operate and combine according to laws and rules of structural linguistics, above all 
the negation. The main law of this field component is the so-called 
{\em Name-of-the-Father}, which triggers the formation of the {\em signifier's chain}. 
This register is the most prominent one in acting on the individual, through the 
intervention of imaginary register.

The {\em imaginary register} is that field component which springs out of the unconscious 
apprehension of one's own bodily image of the child ({\em mirror stage}) on the basis of 
the primary dual relationship of identification with one's own mother. It is the 
basis for the growth, by {\em alterity}, of the Ego agency and the narcissistic pushes, 
when mother, through Name-of-the-Father law, casts the child into the symbolic 
register, naming her or him. 

The {\em real order} is that field component which is defined only in
relation to symbolic and imaginary registers, where there is all that
impossible, unbearable or inexpressible content expelled or rejected by
these latter two registers. The symbolic and imaginary registers,
together with the real order, are in relationship to each other, mostly in
opposition.

A simple example of the action of the three registers is as follows.
This is a good example of Imaginary-Symbolic interconnection. This is
the case of Venice with its renowned carnival. Indeed, this carneval
was instituted around 1090 and as early as that date, many tide 
phenomena flooded Venice. I.e., the unconscious-Imaginary
impregnating Venice meant the coming about of this institution of the
Symbolic to quite popular malcontents due to social status differences
(just featuring the Symbolic), levelling these for instance with masks
which made possible the anonymity, the indistinctness, that are most
typical of such unconscious relationship as the mother-child
relationship of the Imaginary. All the artistic creativity typical of
Venice carnival is due, we can claim, to the irruption of the Imaginary
(tide and flooding) in the Symbolic. This agrees with the well-known
interest of Lacan toward surrealism! In this case, we might claim
further that the fear of death due to the flooding of sea water, just
belonging to the Real, is such that we have a practical example of
Borromean interconnection We observe the Real-Imaginary-Symbolic
relatationship.

\begin{longtable}{|ll|}\hline
Lacan registers  & Persona    \\ \hline
                 & Act 1      \\ \hline
Real              & Queen aware of the letter's content \\
                 & (just belonging to Real register). \\
                 & Inconceivable content of the letter, \\
                 & besides, unknown. \\ \hline
Imaginary         & Queen worried about letter and its \\ 
                 & content. This was then hidden. \\ 
                 & The Queen belongs to this register \\
                 & as she has hidden the letter; this is a \\
                 & behaviour just belonging to \\ 
                 & Imaginary register. \\ \hline
Symbolic         & Minister seizes letter using \\
                 & apparent substitution with own \\
                 & letter. The King, as main signifier \\
                 & giving rise to the symbolic chain to \\
                 & which he is inscribed. \\ \hline
                 & Act 2 \\ \hline
Real             & Police also unsuccessful. \\
                 & Police were on Queen's request. \\
                 & Hence unaware. \\
                 & [Required solution: link between \\
                 & real (the Queen, the only persona to \\
                 & know letter's content) and symbolic \\
                 & (Police interviews because of it \\
                 & being inscribed in a symbolic \\
                 & order).]     \\ \hline 
Imaginary        & This Imaginary register, in which \\
                 & operates Minister (as a robber), \\
                 & then seen as a poet (belonging to \\
                 & imaginary register, the one related \\
                 & to mere creativity and art), thanks \\
                 & to which prefect didn't catch him. \\
                 & Indeed, Poe says too that prefect \\
                 & (belonging to symbolic, i.e., the \\
                 & blind person who does not see the \\
                 & letter) would have caught him if he \\
                 & had had a mathematician's \\
                 & behaviour (symbolic -- see next Act \\
                 & 3), as Minister was both a poet \\
                 & (imaginary) and a mathematician \\
                 & (symbolic). \\ \hline
Symbolic         & Dupin, having his aim disguised, \\
                 & sees probable letter; returns; \\
                 & seizes letter using apparent \\
                 & substitution with own letter. \\
                 & Prefect and Police intervention \\
                 & because of their nature and \\
                 & behaviour which make them \\
                 & belonging to symbolic register. \\ \hline
                 & Act 3 \\ \hline 
Real             & Minister unaware, could be \\
                 & threatened also by this affair. \\
                 & Letter with its content now known \\
                 & to Dupin and which might be \\
                 & revealed. \\ \hline 
Imaginary        & Dupin replacement letter had \\
                 & sinister sentence. Dupin's revenge \\
                 & in regard to Minister, left to be \\
                 & presaged or described or guessed \\
                 & by the Crebillon sentence written \\
                 & by Dupin. \\ \hline 
Symbolic         & Here: the letter, the signifier, in its \\
                 & circuit. \\
                 & [It was/is real; the imaginary was \\
                 & associated with it; \\
                 & symbolic related to apparently \\
                 & similar letters, and \\
                 & also being related to various \\
                 & associated contexts.] \\
                 & Minister seen not as a poet (Act 2) \\ 
                 & but as a mathematician with \\
                 & accordingly a behaviour belonging \\
                 & to symbolic register. \\ \hline 
\caption{Very summarized rendition of the Poe story. Summary of participant 
roles, relative to Lacanian registers.}
\end{longtable}

The main message of this Lacan seminar is to stress the predominace of
symbolic order in constituting human being as such, illustrating this by
means of a Poe story in which Lacan emphasizes how a simple
signifier (the letter, which reifes or materializes, according to Lacan,
the death agency) and its pathway, determines the whole scenes, in
particular, it determines the succession of the three personages
involved there, with their role, each of whom occupies that position just
determined by the letter (signifier) and its movements, which is never
where it is as it is the symbol (in that, signifier) of an absence. On its
turn, the seminar also stands out the imaginary's impregnations owned
by symbolic chain, which mark the unavoidable insistence of the death
drive by means of the compulsion to repeat mechanism.

Lacan underscores the precedence of signifier (letter) on signified
(letter's content), which is besides unknown. What is important is that
there is contained in this letter, whatever is its effective content which,
nevertheless, cannot be revealed (real order) because it is a letter
addressed to the queen who is inscribed into a precise symbolic register
that, as such, surely warrants, in the symbolic chain in which it is
inserted, the symbolic imposing, a priori, of the (unknown) signified
just vehiculated by the (known) signifier. This latter will mould the
disposition of the other various signifiers (personages) along a chain
which will give rise to symbolic order. So, it will be the various
personages of the scenes to be placed along a well-determined chain (of
signifiers) of the symbolic order, and only this chain will be given,
signified, to every subject (personage) so involved: the queen and the
king are placed near real register because of the not revealable content
of the letter (which is ineffable just because it belongs to the real
register); the minister is mainly placed in the imaginary register
(because of his feminine curiosity, intricacy and narcissistic push
which led him to become even a thief), as well as Dupin, but with
touches also with real register, which constantly provides the right fear
to the intrepid actions which they perform; while police and prefect are
located in the symbolic register because they represent the law and are
in search of what belongs to symbolic, i.e., the letter, but without
results because fully immersed in it, without the right amount of
imaginary needs to see. Thus, signified (meaning) springs out only
once the chain of signifiers (words) is established into the symbolic
order. So, the symbolic order with the predominance of signifier on
signified, moulds, in a well-determined chain, the various personages
and their intersubjectivity: there cannot exist a single, isolated
subjectivity but rather an inter-subjectivity provided by the reciprocal
opposition between the elements existing in the chain along which they
are placed, inserted just by symbolic order. Hence, it is the symbolic
order of signifiers that give a precise disposition to the personages in
action in the scenes, giving them too an intersubjectivity which exists
only within this chain that determines them.

The consequences of the (besides unknown) content of the letter belong
to the real register. The chain of signifiers that it emits, belongs to the
symbolic register, while it is the imaginary which glues together the
rings of such a chain, the register that can spring out only thanks the
intervention of a woman (indeed, only the queen knows the real content
of the letter) during the relation child-mother.

In Table 1, there is a useful, very summarized, rendition of the 
Edgar Allan Poe story. It is structured as what we label here as the 
succession of Acts 1, 2, 3. One register will dominate others 
synchronically, i.e. at any given time-point. The symbolic register will 
win out, in that there is a fairly natural progression from the real, to the 
imaginary, thereby resulting in the symbolic. The real register is 
occupied by what the symbolic ejects from reality, and that cannot by 
formalized by language. 

In this first study, the Poe story consists of 321 sentences, and a 
corpus of 1741 words. These words are of length at least 1, all 
punctuation has been removed, and upper case has been set to lower 
case. Then we require a word to be present at least 5 times, and used in 
at least 5 sentences. Next, words in a stopword list were removed. 
These are (definite, indefinite) articles and common parts of verbs, and 
such words (using the {\em tm}, text mining, package in the R software 
package). Single letter words were also deleted (e.g. ``s'' resulting from 
``it's'', or ``d'' resulting from ``didn't'', when the apostrophe here was 
replaced by a blank). Then just nouns alone were selected. There were 
48 nouns at this stage. Some of the 321 sentences became empty. There 
were 213 non-empty sentences, as noted, crossed by 48 nouns. In the 
213 sentences, there were 424 occurrences of these words.

The sentence set, characterized by words used, endowed with the 
chi squared metric is mapped, using Correspondence Analysis, into a 
Euclidean metric-endowed factor space. In order just to retain the most 
salient information from this semantic, factor space, we use the 
topmost 5 axes or factors. These 5 axes account for 17.75\% of the 
inertia of the sentences cloud, or identically of the nouns cloud. Figure 
1 displays the hierarchical clustering of the sentences, that are in their 
5-dimensional semantic or factor space embedding. The complete link 
agglomerative clustering criterion permits adherence to the sequential 
order of the sentences (Murtagh, Ganz and McKie, 2009; B\'ecue-Bertaut 
et al., 2014, Legendre and Legendre, 2012).

\begin{figure}
\centering
\includegraphics[width=\textwidth]{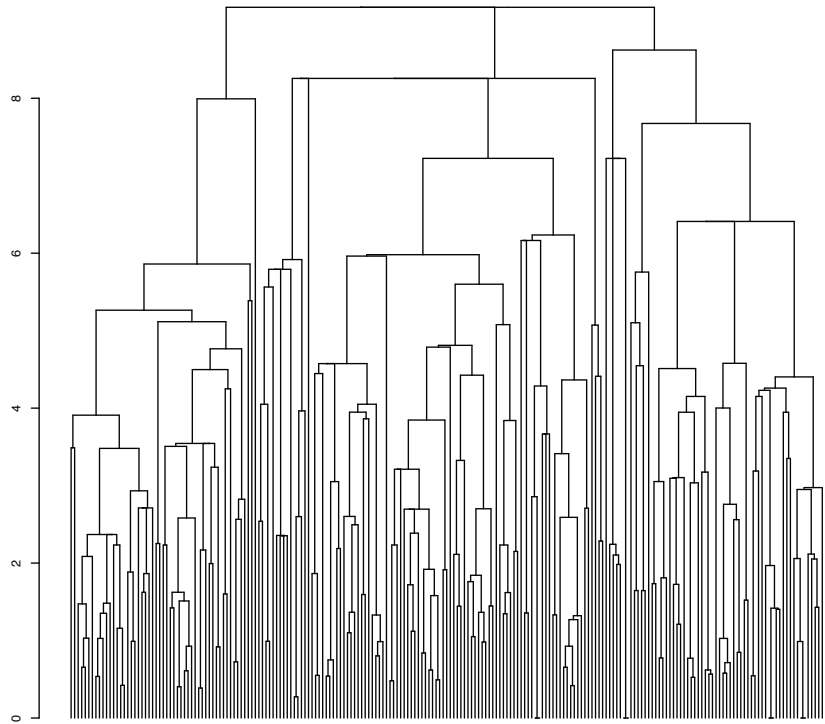}
\caption{Hierarchical clustering of sentence by sentence, based on the story's
sequential structure.  There are 213 sentences here, being the terminal (or leaf)
nodes, ordered from left to right.  Each sentence contains some occurrences from
the corpus of 48 nouns that are used. The vertical axis, for such a dendrogram,
records the cluster criterion agglomeration values and levels.}
\end{figure}

To follow our template of three acts, we take the three largest 
clusters. In the dendrogram in Figure 1 we therefore have the partition, 
containing three clusters, close to the root node. These clusters relate to 
sentences 1 to 53, sentences 54 to 151, and sentences 152 to 213. These 
are to be now our acts 1, 2, 3, following the template set out 
descriptively in Table 1. The number of sentences in each of these acts 
is, respectively, 53, 98, 62. Next for analysis, we create a table crossing 
3 acts by the noun set of 48 nouns. 

The complete factor space mapping is just in 2 dimensions.  We may
just note the visualization benefits that follow the relating of nouns to
what we term the acts, rather than the individual sentences. Figure 2
displays the words that have the highest contribution to the inertia of
this plane. To see the relationship between the words that are close to
the origin, thus essential to the whole of the narrative line, to all acts,
Figure 3 displays the region of the plane that is close to the origin. We
see ``letter'' and other words.

\begin{figure}
\centering
\includegraphics[width=\textwidth]{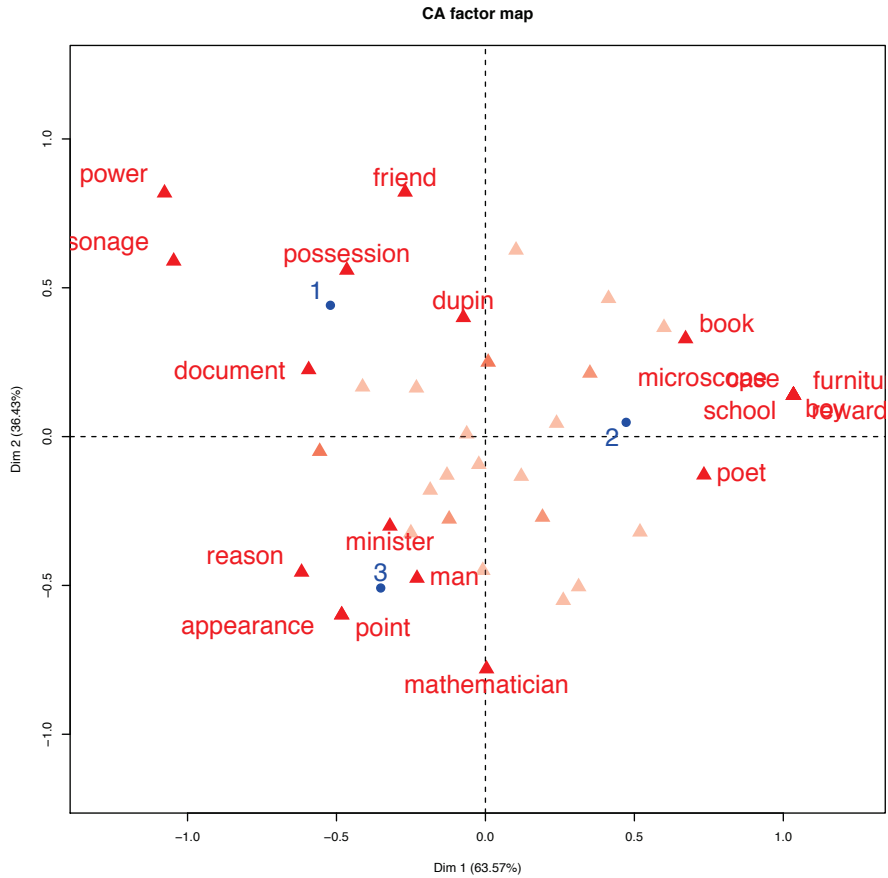}
\caption{Correspondence Analysis, top contributing 20 words.  Words with high
contribution, somewhat overlapping in this display, with projections on the positive
factor 1 are: case, microscope, school, reward, boy, furniture; and book, poet.
Also displayed are the three acts, 1, 2, 3.}
\end{figure}

\begin{figure}
\centering
\includegraphics[width=\textwidth]{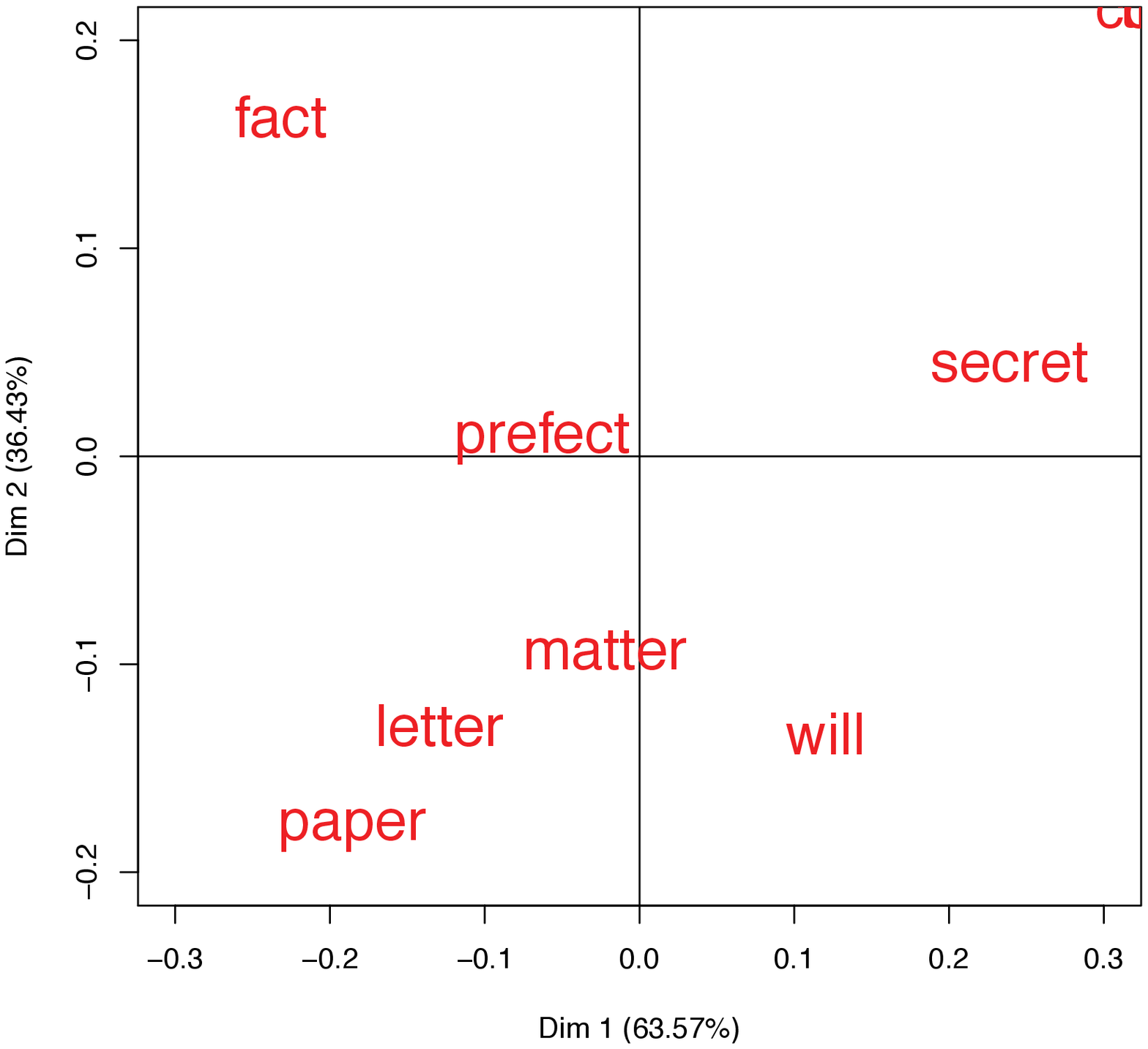}
\caption{From Figure 2, here are shown the words that are near the
origin.}
\end{figure}

We can try to investigate the internal structure of our ``template'' 
acts. Figure 4 displays the hierarchy (using the appropriate 
agglomerative criterion of Ward's minimum variance) constructed in 
the 5-axis or 5-factor embedding of this data. From left to right here, 
the three clusters resulting from the dendrogram cutting, or slicing into 
a partition, as displayed, correspond mapping-wise to act 2, act 3, act 1. 
Cf.\ what is displayed in Figure 2.

\begin{figure}
\centering
\includegraphics[clip,trim={0 4cm 0 4cm},width=\textwidth]{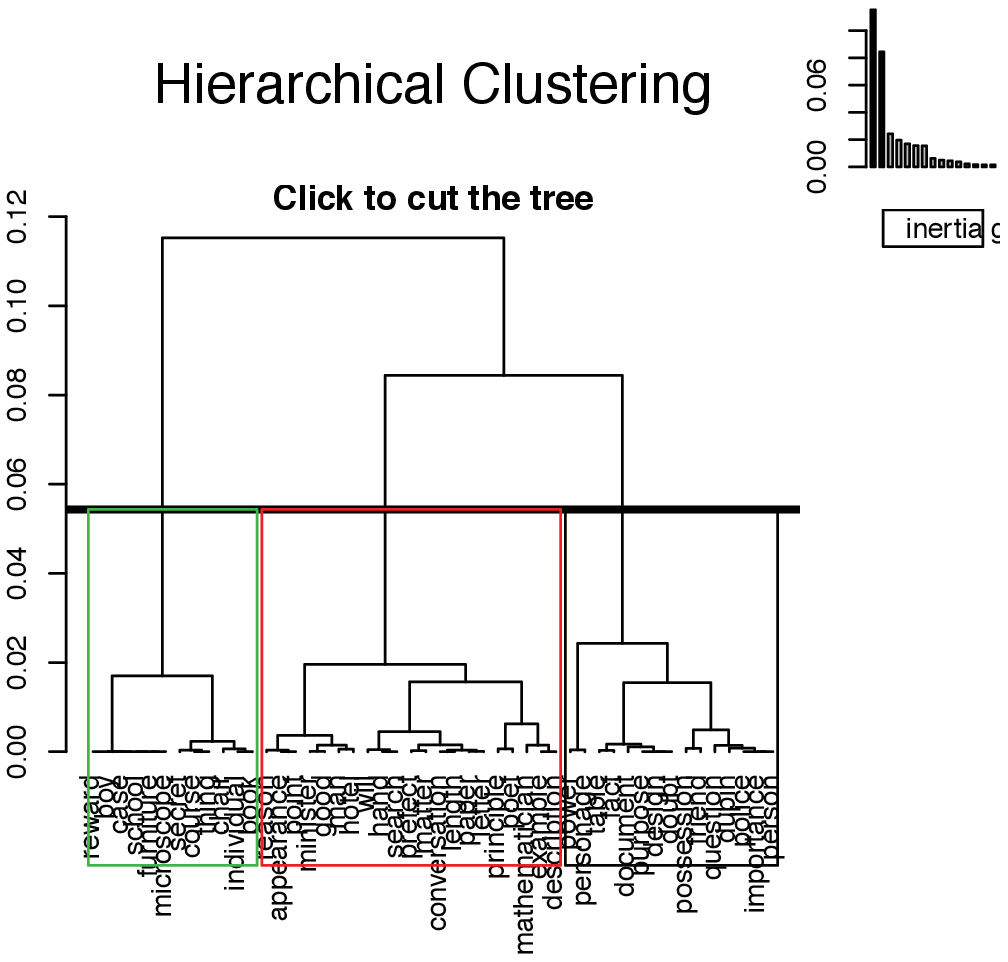}
\caption{Hierarchical clustering of the word set, from the 2-axes, correspondence
factor space, semantic mapping of this data. Here the hierarchy structure is
visually displayed.}
\end{figure}

With the perspective of Lacan’s registers we could look at a set of 
three clusters in each of these acts. Let us look at the leftmost cluster 
here. We can read off the following three clusters: first cluster, 
``reward, boy, case, school, furniture, microscope''; second cluster, 
``secret, course, thing''; third cluster, ``chair, individual, book''. This has 
just been reading off three fairly clearly determined clusters. Of course 
we can see that the second cluster and the third cluster are merged 
fairly early on in this agglomerations. 

Rather than attempting to relate these clusters with Lacan's 
registers, let us instead just draw the following conclusion. In this 
first study, it has been shown how a template of segmentation can be easily 
considered. So the diachrony can be investigated. In our opinion, the 
retained words consisting of nouns are a good way to focus our study, 
and also while we succeeded well in imposing our template of the 
segmentation of the Poe story into three acts. However while they 
certainly lead to interesting perspectives, for general-purpose use of 
this methodology, it would be preferable to allow for a somewhat more 
open perspective on the data. This we do next, analysis of diachrony 
and of synchrony, both newly investigated. 

\section{Visualizing Lacanian Registers in the Narrative Flow}

\subsection{Lacanian Framework}

Fundamental aspects of Lacanian methodology encompass the following (Richardson, 1985). 
\begin{itemize}

\item Metonymy, e.g.\ the name of the cause is used for denoting the 
effect or the object; it is associated with, and expressed by, 
diachronicity. Diachronicity horizontally combines patterns into 
metonymy. Finally, metonymy is to be associated with 
(Freudian) unconscious displacement, shifting and moving, 
under the pushes or drives of desire, the various signifiers, 
without an end but rather aimed always at seeking the lost object 
(lacking for the human) which escapes every signification. 

Metaphor, e.g.\ a word or term is replaced with another similar or 
analogous term; through selection, metaphor is enabled by 
synchronicity. Synchronicity vertically selects patterns into 
metaphor. Finally, metaphor corresponds to (Freudian) 
unconscious condensation, which disguises and upsets meanings, 
until reaching the deepest unconscious levels.

Metonymy and metaphor are the two main (Freudian) pathways 
of semantic action.

\item Signifiers will constitute the symbolic register. These signifiers 
combine like the basic structural elements of a language. The 
signifiers of the symbolic register undergo the rules of metonymy 
and metaphor.

For Lacan, the signifier dominates the signified, and not vice 
versa (as for Ferdinand de Saussure), through certain structural 
rules (similar to the linguistic ones) in which the former 
(signifiers) link together to give rise to signifier chains. Signifier 
chains are diachronic combination of signifiers synchronically 
selected, in which the signifiers follow each other oppositionally, like the words of a phrase.

Indeed, (synchronic) selection includes the case where a signifier 
excludes another one but remains in relationship with that other 
signifier, at least negatively, according to Aristotelian logic. 
These signifier chains will acquire then a conscious meaning 
following usual grammatical rules.

\item The Imaginary is a register complementary to the Symbolic one. 
Generally, it is the realm of images and of the sensible 
representations (mostly, the visual ones) which mark our own 
lived experience. Imaginary fantasies and representations (thing 
representations) belong to the imaginary register as well, which 
will prepare the ground for the subsequent word representation.

\item The Real is not reality as this is usually meant, that is to say, the 
world of everyday experience, which is already characterized by 
images and symbolic language, but it rather deals with the 
primary, rough experience of what is still not symbolized or 
imagined, with the impossible, that is to say, what is impossible 
to inscribe in every symbolic system, or however represented in 
any possible imaging form.
\end{itemize}

\subsection{Narrative Flows}

All discourses, happenings, history, etc.\ are narratives, with one or 
more, and often many, narrative flows. In the narrative, there are 
various chronologies that may be investigated as sub-narratives. These 
include the sequence resulting from: (i) sections, (ii) speaker or agent, 
(iii) time or date or location, (iv) statistical segmentation into sections. 
The latter may be through syntax and style based clustering since tool 
words (function words) predominate. To the above can be added: (v) 
sentences, (vi) paragraphs. All this, is the result of the diachronic 
nature of the discourse, which, therefore, is explainable through 
Lacanian theory. In particular, Lacan points out that, in the symbolic 
register, the diachronic selection axis of discourse is closely related 
with synchronic combination of signifiers which gives rise to the 
diachronic meaning, or signified, of the discourse. These combination 
and selection processes, taking place in the symbolic register, are 
greatly influenced by the real register and, especially, by the imaginary 
register. These latter both push on the former.

We seek the most enlightening or the most illustrative of these 
narrative flows. By enlightening, we intend: seeking or determining 
specific outcomes. By illustrative, we intend: detecting or observing 
dialectical movement, or Aristotelian logic, or unconscious mind processes.

We are most interested in (i) metaphor, being an indicator of 
unconscious mind processes, for its synchronic nature, and (ii) 
metonymy, i.e.\ a term indicating diachronic employment (or use), that 
can be, therefore, transfer and handover.

Following the mapping of the text story into a semantic space, in 
regard to combinations of signifiers according to Lacan, for (i) 
metaphor, due to its synchronic nature, we use clustering. While, for 
(ii) metonymy, due to its diachronic nature, we use sequence 
constrained, i.e.\ chronologically constrained clustering. In (i) our aim is 
close association, expressed by highly compact clusters, while in (ii), 
we may consider varied chronological flows.

Our semantic analysis starting point is the set of all interrelations 
between narrative flow segments, belonging to the diachronic selection 
axis, and the words selected and retained, belonging to synchronic 
combination axis. We have that: ``One terms the distribution of a word 
the set of its possible environments'' (Benz\'ecri, 1982).

\subsection{Text Narrative Analysis: Initial Processing Stages}

The Poe story, in our text formatting, consists of 321 sentences, 
arranged as 123 paragraphs. As noted above, paragraph here is defined 
as text segments that are separated by blank lines. That includes vocal 
expressions, perhaps with some additional explanatory text, and also it 
may be noted that a few of the vocal expressions can be quite short. 
Nonetheless it is clearly the case that the paragraphs form useful text, 
and narrative, segments. 

Next we also considered a segmentation into 8 sections, based on a 
reading of the Poe story. The introduction part of the story had 19 
paragraphs. The initial outlining of the essential story, relating to the 
purloined letter, told by the Prefect with dialogue elements from the 
narrator and from Dupin, constituted section 2, with 26 paragraphs. 
Section 3, with 28 paragraphs recounted the Prefect's search of the 
Minister’s hotel room. Section 4, with 14 paragraphs, takes place one 
month later, detailing the revelation that Dupin could provide the letter 
to the Prefect. Then section 5, with 6 paragraphs, starts off the 
background explanation by Dupin to the narrator. Section 6, with 16 
paragraphs, continues in great detail as Dupin provides explanation to 
the narrator. Section 7, with 8 paragraphs, is the core of the storyline, 
where Dupin explains how he found the letter, how this was verified by 
him, and how he took hold of it in the following morning, putting what 
is referred to as a facsimile in its place. Finally section 8, with 6 
paragraphs, is the explanation of, and justification for, the replacement 
of the letter by a facsimile. 

Because of the consolidated and integrated description, with 
motivation and explanation, Dupin's explanation of all of this, in 
sections 5, 6, 7, 8, may be additionally considered in our analysis. 
We have just noted the paragraphs that correspond to these sections. 
Section 5 begins with sentence 172 (in the set of 321 sentences). So the 
Dupin explanatory sub-narrative, in dialogue with the narrator, 
embraces sentences 172 to 321, that is, paragraphs 88 to 123. So the 
Dupin sub-narrative here comprises 151 sentences, that are in 36 
paragraphs. 

Our next step in data preprocessing is to select the word corpus that 
will be used. This starts with removal of all punctuation, numeric 
characters, and the setting of upper case to lower case. 

It is reasonable here to remove tool words, also referred to as 
function words. In Murtagh (2005, chapter 5), and in Murtagh and 
Ganz (2015), the case is made for these function words in mapping 
emotional narrative or stylistics (e.g.\ to determine authorship), but 
these are not of direct and immediate relevance here. Instead, as 
outlined in section 2.1, metaphor and metonymy are the forensic 
indicators, or perhaps even the forensic highlights, for us. 

Sufficient usage of the word in the storyline is important. While 
very clearly the case that one-off (isolated, unique) use of a word can 
be very revealing, nonetheless we leave such an investigation to an 
alternative comparative study of storyline texts. Sufficiently frequent 
word usage both supports comparability between the text units we are 
studying, and also permits the focus of the analysis to be on 
inter-relationships, and not on uniqueness of word usage. Therefore we 
require the following for our word corpus: that a word be used at least 3 
times in the overall storyline, and that this word be used in at least 3 
of the text units (sentence, paragraph, section) that we are dealing with. 

For the 321 sentences, we start with 1742 words. There are, in total, 
7089 occurrences of these words. Then, having removed stop words, 
and requiring that a word appear in 3 sentences and be used at least 3 
times, we find that our 321 sentences are characterized by 276 words. 
There are 1546 occurrences, in total, of the corpus of 276 words. 

For the paragraphs, proceeding along the same lines, the 123-paragraph 
set is characterized by the 276 word set, and there are, as for 
the sentence set, 1546 occurrences, in total, of the corpus of 276 words.

For the sections, once again proceeding along the same lines, the 
8-section set is characterized by the 276 word set, and, again clearly, 
there are 1546 occurrences, in total, of the corpus of 276 words. 

This data preprocessing and selection is carried out for the following objectives.

Firstly, we will have one or more levels of text (hence, storyline) 
unit aggregation so that the principal factor space axes account for most 
of the information content. (Were it the case of having rare words in 
the analysis, then axes would be formed in the factor space to account 
for them.) We recall that for $n$ text units, characterized by $m$ terms, the 
factor space dimensionality will be min($n-1$ , $m-1$). This first point 
relates to the use of paragraphs and sections. (Let us note that in 
B\'ecue-Bertaut et al.\ (2014), where the flow and evolution of narrative is the 
aim, our aim is a little different here, because the text units that 
encompass the most basic text units, the sentences, can be themselves 
interpretable. Cf., e.g. vocal expression on a theme being all in one 
paragraph.)

Secondly, our selection of words directly impacts the interpretation of the data. 

\subsection{A Preliminary Visualization of the Narrative Structure}

We have here the successive sentences characterized by their 
constituent words, from the retained corpus. We firstly map the cloud 
of sentences, 321 sentence cloud in a 276-dimensional word set space, 
into a Correspondence Analysis factor space. Since the word set has 
been reduced from the original set of 1742 words, some sentences 
become empty. Non-empty sentences account for 310 of these 321 
sentences. In Figure 5 we required words to be at least 5 characters 
long. This led to a corpus of 205 words, with 293 sentences not 
becoming empty. 

\begin{figure}
\centering
\includegraphics[width=14cm]{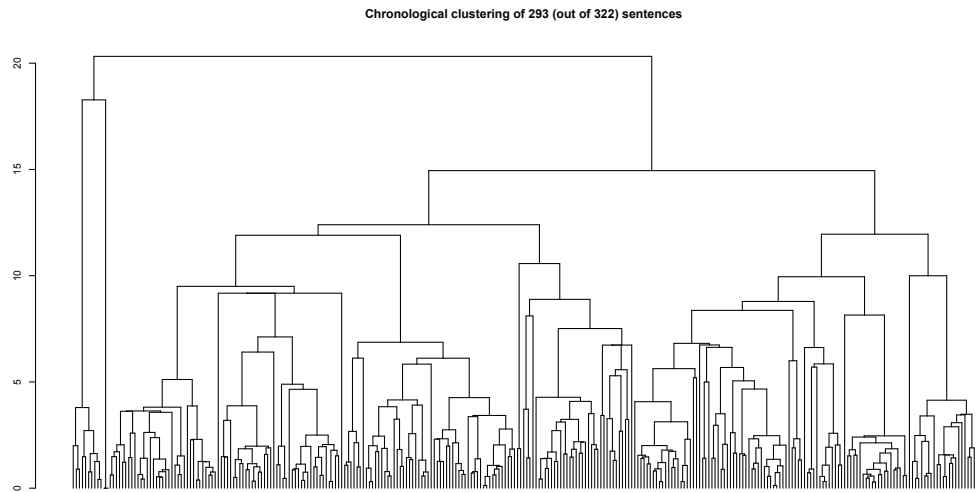}
\caption{Contiguity-constrained, where contiguity is the chronology or
timeline, hierarchical clustering of the 321 sentences. These sentences are
characterized by their word set (1087 occurrences of 205 words). This hierarchy is
constructed in the factor space, of dimension 5, that is endowed with the
Euclidean metric. Due to the reduced word entailing that some sentences become
empty, the number of sentences in the correspondence factor analysis was 293 (from
the 321). Here the dendrogram structure is displayed.}
\end{figure}

In Figure 5, sentences 11 and 12 are merged very early in the 
sequence of agglomerations, and these sentences are found to be quite 
exceptional. They are as follows: Narrator: ``Nothing more in the 
assassination way I hope?'', Prefect: ``Oh, no; nothing of that nature.'' 
The two large clusters that are merged at the 3rd last agglomeration 
level have the last sentence of the first large cluster, and the first 
sentence of the second large cluster as follows. Sentence 182: ``But he 
perpetually errs by being too deep or too shallow for the matter in 
hand; and many a school-boy is a better reasoner than he.'' Sentence 
193: ``I knew one about eight years of age, whose success at guessing 
in the game of ‘even and odd’ attracted universal admiration.'' This is 
early in what has been taken as the Dupin explanatory section of the 
narrative. The second very large cluster constitutes the major part of 
this Dupin explanatory part of the storyline. 

In B\'ecue-Bertaut et al.\ (2014), it is described how the text units, 
taking account of the chronological order, can be statistically assessed 
(using a permutation-based statistical significance testing) at each 
agglomeration, for the agglomeration to be based on a pair of 
homogeneous clusters. This allows derivation of a partition of the 
set of text units. Since the chronological, hence contiguity, constraint applies, 
this partition is a statistically defined segmentation of the text units. In 
this particular work, we prefer to use paragraphs and sections, as 
described above, in view of their interpretability. 

\subsection{Visualization of Lacanian Registers from Semantic Analysis of Chronology Using Storyline Segments}

In this section, we are most concerned with diachrony, or the evolution 
of the narrative. For this, we find a correspondence – what we may 
refer to as homology, in the sense of Bourdieu-related geometric data 
analysis -- between a pattern that we uncover in the data, and Lacan's 
registers, viz.\ the Real, the Imaginary and the Symbolic. In the 
storyline here, we find an evolution, or narrative trajectory, between 
these registers. Lacan’s registers are of value to us as an interpretive 
viewpoint. It has been noted above in section 4.1, how both synchrony 
and diachrony of the semantics of the storyline narrative are of 
importance here. As noted also, we can determine statistically a 
segmentation of the narrative. This is achieved through first mapping 
the narrative into the semantic factor space, taking account of all 
interrelationships of narrative text units and the words and terms that 
are associated with these text units. For interpretation, we prefer, see 
section 4.3, to use what we have selected as natural segments in the narrative text.
 
The cumulative percentages of inertia associated with factors 1 to 7 
are as follows: 20.2, 38.1, 54.2, 68.8, 81.5, 92.9, 100. The principal 
factor plane is displayed in Figures 6, 7. The chronological trajectory is 
to be seen in the first of these figures. The second figure has a 
triangular pattern, that is a display of the narrative, with reference to 
the chronology of the narrative. Usually with such a triangular pattern, 
we look especially towards the apexes in order to understand it. Figure 
7 shows the most important words.

\begin{figure}
\centering
\includegraphics[width=\textwidth]{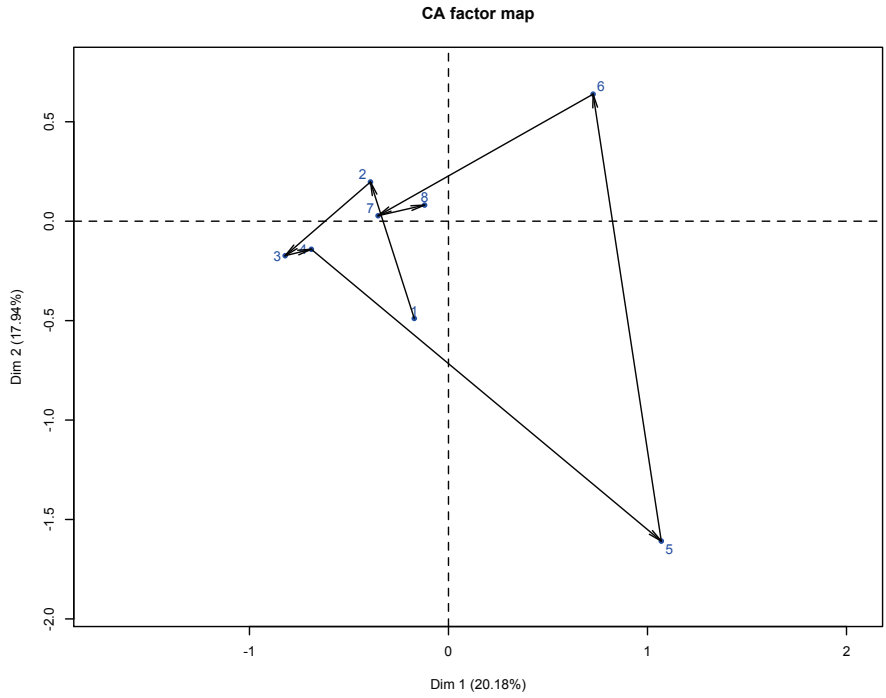}
\caption{Principal factor plane of the 8 story segments crossed by
the 1546 occurrences from the selected 276-word corpus.  Arrows link
the successive segments, numbered 1 to 8.}
\end{figure}

\begin{figure}
\centering
\includegraphics[width=14cm]{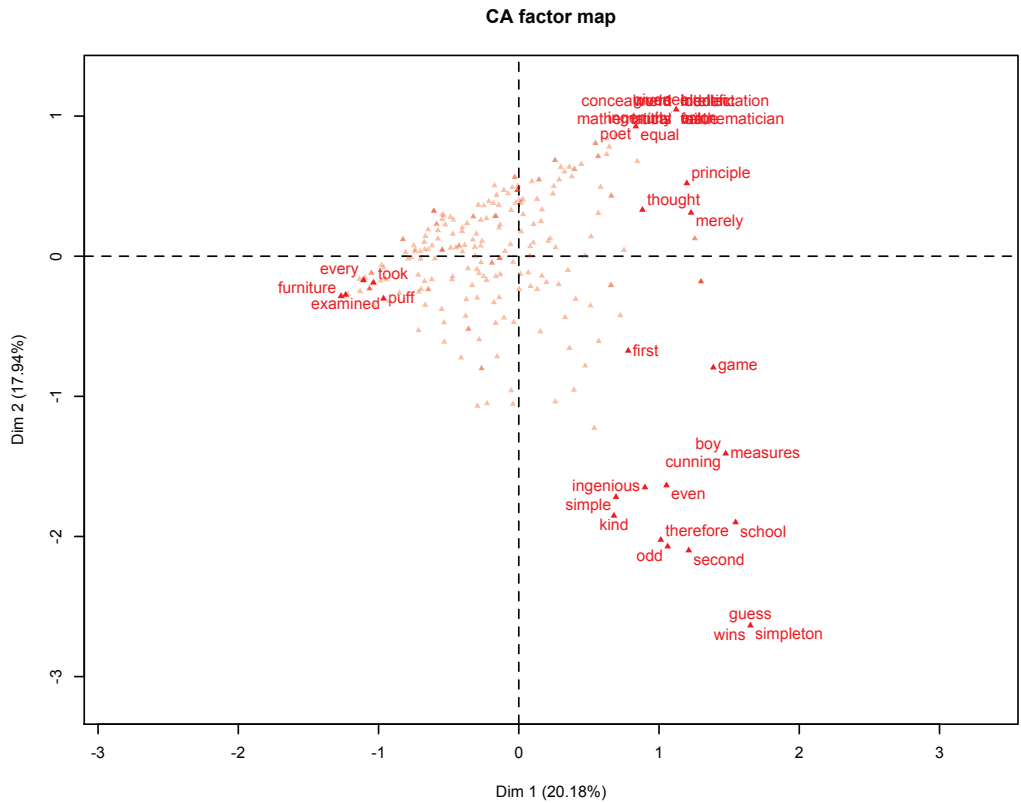}
\caption{Displayed here are the 40 words that most
contribute to the inertia of these factors, factors 1 and 2. In the upper right
(beyond {\tt equal, poet}), terms are:
{\tt mathematician, world, value, truths, see, mathematical, intellect,
fail, error, ingenuity, identification, hidden, given, concealment, reason, hand}.}
\end{figure}

From this display, taking the figures 6, 7 (not overlaid in the same 
figure, to make the displays clearer), we can conclude in this way: 
segments 1, 2, 3, 4 are gathering facts and impressions from the Real; 
segment 5 advances into the Imaginary; segment 6 expresses this in a 
Symbolic way; and that allows a consolidated, integrated, ``overall picture'', 
core of segments 7, 8.

In Figures 8 and 9, factors 3 and 4 are displayed. If this viewpoint 
expressed above is acceptable, namely that segments 5 and 6 comprise 
the move towards the Imaginary, then towards the Symbolic, then we 
can draw this perspective: that the effect of these two segments in the 
overall narrative is to take such segments as segments 3 and 4, 
operating in the Real, then work through the Imaginary and Symbolic 
discussion, and arrive then, as a consequence, at the final, terminal and 
more conclusive segments, segments 7 and 8. In a way, we are drawing 
the conclusion, from this particular storyline, as to how the Imaginary 
and the Symbolic serve to be taken into (and become part of) the Real, 
or how the Symbolic emerges from the Real and the Imaginary. 

\begin{figure}
\centering
\includegraphics[width=\textwidth]{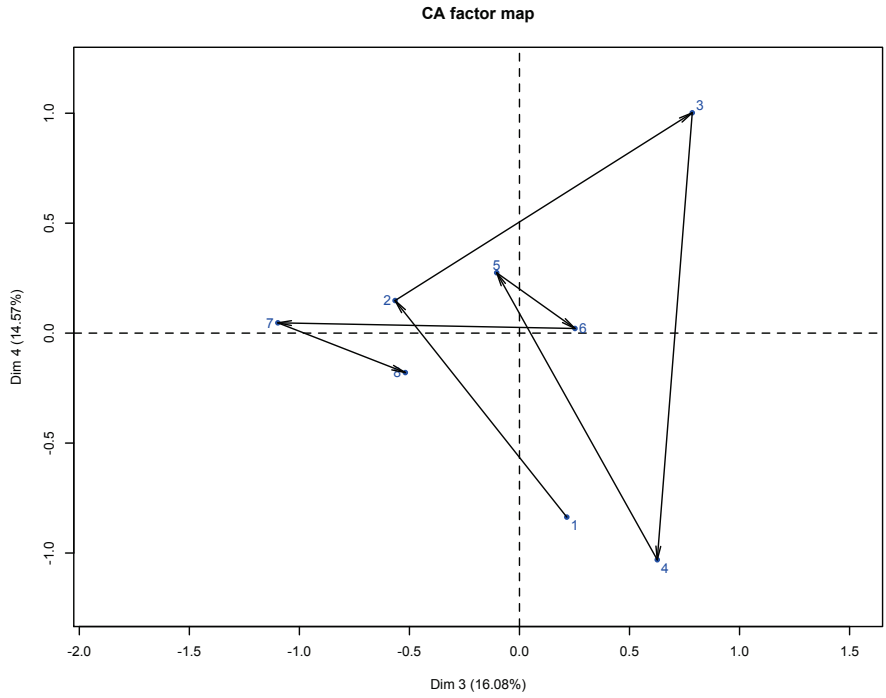}
\caption{Plane of factors 3, 4, displaying the 8 story segments,
with arrows linking the successive segments, numbered 1 to 8.}
\end{figure}

\begin{figure}
\includegraphics[width=14cm]{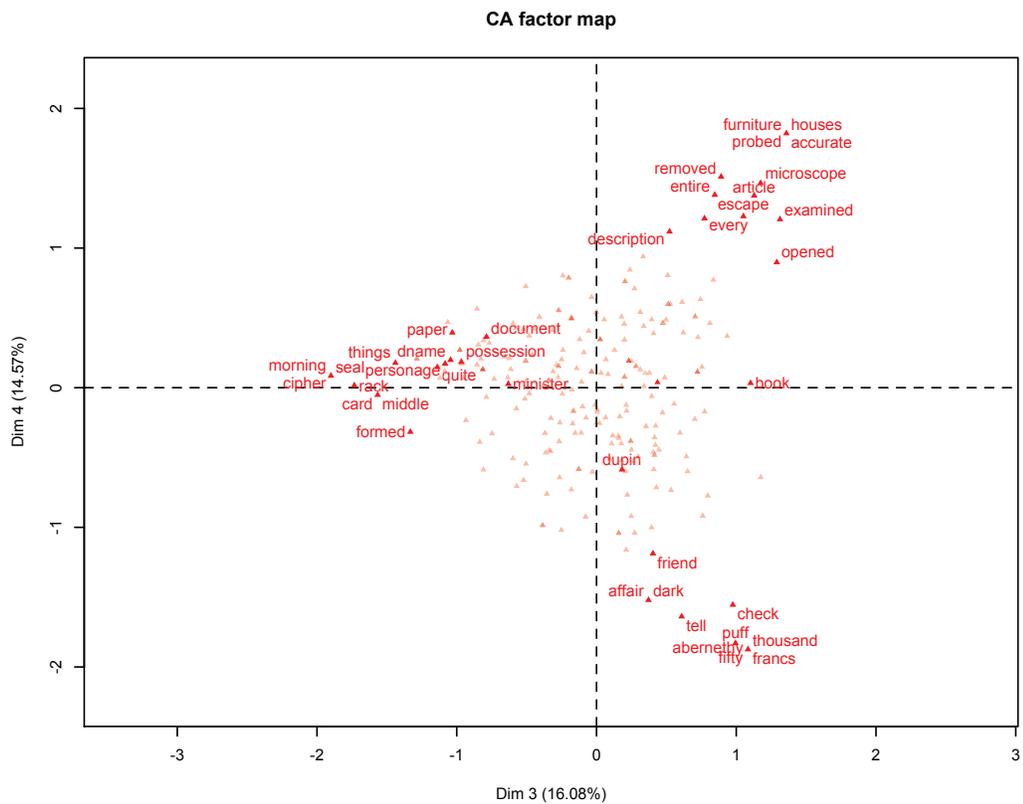}
\caption{The 40 words that most
contribute to the inertia of these factors, factors 3 and 4.  The word
{\tt dname} is a rewritten form of ``D--'', i.e.\ Minister D.  In the text,
there is discussed: the Real, left here; the Imaginary, upper right; and the
Symbolic, lower right.}
\end{figure}  

Moving on now to the third and fourth factor plane, Figure 9, there 
is a more interesting and selective perspective, given our interest in 
having an informative visualization of Lacanian registers. We propose 
the following perspective on this figure, Figure 9. Take the words on 
the left, negative half axis of factor 3, as pertaining to a Real register. 
Therefore, mostly, they betoken the unknown or the unknowable. Next, 
take many of the words displayed in the upper right quadrant as 
associated with the Imaginary. This includes ``furniture'', ``houses'', 
``microscope'' and so on. This is how we can imagine problem-solving. 
Thirdly, and finally, take many of the words in the lower right quadrant 
as betokening the Symbolic. What we have here is money, payment. In 
other words, in a practical setting here, the problem solving is 
associated with the symbolic value of money. 

We conclude that Lacan's registers have been of major benefit in 
providing semantic-related understanding of the essential pattern that 
we determined in the narrative chronology. Such homology of semantic 
structure, i.e. morphology of narrative, is to be sought in any domain, 
such as the Poe story here, that can be modelled through Lacan's 
registers. 

\section{Textual Data Mining as a Basis for Discovery Metaphor and Metonymy}

\subsection{Contextual, Semantic Clustering}

In order to provide a basis for metaphor and metonymy finding, we will 
use the dominant words in clusters that we determine. For general data 
mining, a particular selection of words is used. 

In summary, from section 3, we start with the 1741 words derived 
from the 321 sentences in the Poe story. Firstly, in order to exclude 
anomalous words, because of their exceptionality, we require that a 
word is used in at least 5 sentences, and that the word is used at least 5 
times. This will also remove the small number of French words used in 
the Poe story. Secondly we exclude English stopwords. That resulted in 
127 words being retained. In this general data mining framework, 
thirdly, we decided to retain only nouns. That resulted in 48 words 
being retained. Some sentences become empty through removal of their 
words. This left us with 213 non-empty sentences, i.e.\ 213 sentences 
crossed by 48 words. In addition, to focus our data mining, we used the 
three acts in the Poe story, as defined in section 3. That aggregated the 
sentences comprising these three successive parts of the story. So we 
use two data sets, 213 sentences crossed by 48 words, and 3 acts 
crossed by 48 words.

Carrying out a hierarchical clustering on the full Correspondence 
Analysis ensures that the Euclidean metric endowed space is fully 
appropriate to have clustering carried out, using the minimal variance 
agglomerative criterion (i.e. minimal change in inertia, or variance 
since all masses are unity). The full factor space dimensionality is used 
so that there is no loss of information through reduction in 
dimensionality. Such is not always the best approach because it could 
be argued that the principal factors represent the essential 
interpretational information. 

From the 213 sentences ×48 retained set of nouns, the minimum 
variance (or Ward criterion) hierarchical clustering gave a partition into 
10 clusters, using the greatest change in variance. For the 3 acts $\times$ 48 
retained set of nouns, the same criterion gave a partition into 3 clusters.

For the first of these, there are the following non-singleton clusters: cluster 1, 
{\tt boy, school}; cluster 5, {\tt individual, microscope, doubt}; cluster 7, 
{\tt letter, prefect, dupin, minister, document}; cluster 8, {\tt paper, power, secret}; cluster 9, 
{\tt poet, mathematician}; and cluster 10, {\tt design, reason}. Very close 
semantic similarities are clear here. We may consider poet being a 
metaphor for mathematician, and vice versa. While letter is strongly 
related to paper, document, it is also metaphorically related to power 
and secret. 

For the acts crossed by words data, there are the following clusters: 
cluster 1, {\tt table, fact, document, possession, dupin}; cluster 2, {\tt conversation, length, hotel, good, paper}; 
and cluster 3, {\tt book, individual, boy, case, furniture}. We interpret this 
output as overly concentrated, that it can be considered in relation to 
Lacanian registers, but that it is of less directly interpretable value 
compared to the previous output, described in the previous paragraph. 

Another approach to addressing the discovery of metaphor, and 
related interpretable outcomes, is to carry out the clustering -- 
hierarchic clustering, followed here by partitioning -- on the sentences, 
and then to investigate the words that are statistically significant for the 
clusters that are found. Hypothesis testing is carried out using the v-test 
(Husson et al., 2011). For the acts crossed by words data, there is not 
great statistical significance. For the sentences crossed by words data, a 
most interesting set of three clusters, in the greatest change in variance 
partition, is obtained. Cluster 3 (with arbitrary numerical labelling of 
clustering) has the words {\tt reason} and {\tt design}. Cluster 2 has the 
words {\tt poet} and {\tt mathematician}. Finally, cluster 1 has all of these 
words: {\tt reason} and {\tt design}, and {\tt poet} and {\tt mathematician}. 
Very interestingly, we find here that poet and mathematician are 
metaphorically related through their involvement in reason and in 
design. We consider that this also provides for metonymy. 

We may with to look for pointers towards the triad that defines a 
metaphor (e.g., poet, mathematician, and reason; poet, mathematician, 
and design). Consider how Ricoeur (1977, p. 276) conceptualized this: 
``We arrive at metaphor in the midst of examples where it is said, for 
instance, that a certain picture that possesses the colour grey expresses 
sadness. In other words, metaphor concerns an inverted operation of 
reference plus an operation of transference. Close attention must be 
paid, therefore, to this series -- reversed reference, exemplification, 
(literal) possession of a predicate, expression as metaphorical 
possession of non-verbal predicates (e.g.\ a sad colour).'' Thus in brief, 
we may consider here that x = picture, y = grey, z = sadness, and we 
have the proximity of x and y that we may view as comprising the 
apexes of the base of an isosceles triangle. A triangle that is isosceles 
with small base is the defining property of an ultrametric topology (i.e.\ 
representing a tree or hierarchical relationship). Such an ultrametric 
relationship expresses unconscious reason, cf. Murtagh (2012a, 2012b, 
2014). 

From looking at the close semantic (i.e., based on the semantic 
factor space embedding) association of cluster members, we have 
pointers to what could play the role of metaphor, being locally and 
temporally, contextualized synonyms. This is together with what could, 
over a time line, play the role of metonymy. 

A final issue addressed now is in regard to metonymy. Aspects of 
the imaginary and symbolic are potentially of relevance, including the 
poet and mathematician referred to in the purloined letter case, and 
symbolic rationalisation from the school boy with his marbles. 
Essentially, relationships are to be determined and discovered in the 
semantic factor space. They may be then further assessed relative to the 
original data. As an illustration of this, consider a selection of words 
retained, crossed by what we are referring to as acts in the Poe story, 
Table 2.

\begin{table}
\caption{The acts are the successive major segments of the Poe story.  From the 48 words
retained here, the frequency of occurrence data is shown for a selection of 11 words.}
\begin{tabular}{|r|rrrrrrrrrrr|}\hline
Act  & letter & dupin & minister & police & power & prefect & question & reason & reward & search & secret \\ \hline
1 &  8  &  13  & 5 & 2 & 5 & 8 & 3 & 2 & 0 & 1 & 1 \\
2 &  13 &  15  & 6 & 3 & 0 & 13 & 4 & 1 & 7 & 5 & 3 \\
3 &  11 &   4  & 9 & 1 & 1 & 8  & 0 & 4 & 0 & 3 & 1 \\ \hline
\end{tabular}
\end{table}

\subsection{Further Exploration of Statistically Significant Word Associations}

For close associations of words leading to either metaphor or metonymy, we adopt the following 
principles. Firstly, we seek such associations from the data, and we do not impose an a priori 
statistically-based probabilistic model or other prespecified criterion. Secondly we want to 
have such associations contextualized. The latter is for the seeking of associations to be in 
semantically-defined clusters. 

We also investigated the chronology based on the following: the sequence of sentences; the 
sequence of paragraphs, i.e. text segments, that were mostly either a continuous speech segment, 
or relating to an individual; a set of eight sections covering the entire story that was manually 
segmented, approximately in line with the timeline; and four statistical segmentations of the 
storyline based on combinatorial probabilistic significance levels. Successive segmentation of 
the storyline was, respectively, with the following numbers of segments: 321, 123, 8, 46, 26, 13, 11. 
It was found that these sequences were weakly correlated with the factors. As supplementary elements 
on the factor space planar projection, they were very close to the origin. We conclude that there is 
not much that carries chronological meaning in this story. That is on the global or overall level. 
Word associations or sequences (that could play a role in metaphor or in metonymy formation) are a 
different issue. 

Based on the Correspondence Analysis factor space mapping, endowed with the Euclidean distance, 
the clustering of sentences and also of words was investigated. Although distinct in regard to the 
basis for the clustering, while of course using the minimum variance -- hence inertia in the 
Euclidean-endowed factor space -- agglomerative criterion, the outcomes implicitly share the 
5-dimensional (used just by default as a small set of factors) input. 

It has already been noted how factor 1 counterposes the specifics of investigation to the 
ancillary small sub-narratives, relating to mathematical thinking analogies (upper right quadrant) 
and to the marble-playing schoolchild motivation and decision-making analogies (lower right quadrant). 

The 5-class partition obtained allows us to look closely at some of the clusters. These clusters 
are of cardinalities, for the words: 10, 218, 5, 19, 24, and for the sentences: 8, 258, 6, 21, 17. 
They are in sequence of their mean value projections, from left to right on the first axis. 

Let us look at low level partitions in the dendrogram in order to select small cardinality, very 
compact clusters. Following Husson et al. (2011, p. 151) we can use the v-test of association of 
the category presence values relative to the mean value of that variable. This allows for a null 
hypothesis test of ``the average ... for [the] category ... is equal to the general average'', ``in 
other words, [the] variable does not characterise [the] category ... and can therefore calculate 
a p-value''. A p-value not far from zero indicated rejection of that null hypothesis. That is to 
say, a p-value near zero indicates that the variable emphatically does characterise the category. 

When we look at an 11-class partition we find classes 1 and 2 consisting of: 

\begin{verbatim}
Class 1: 
            p.value of H0 using v.test
puff        1.188185e-13
abernethy   1.097031e-03

Class 2: 
            p.value for v.test
probed      1.693836e-06
looked      1.758749e-03
\end{verbatim}

Class 5 with the following words, with p-values of the v-test less than 0.05 (therefore 
rejecting the null hypothesis here at the 95\% significance level): {\tt letter}, {\tt man}, 
{\tt ordinary}, {\tt gname}, {\tt reward}, {\tt asked} (Here {\tt gname} is the Prefect. 
There is for example the following in the 
Poe text: ``Monsieur G -- -- , the Prefect of the Parisian police.''). In this 11-cluster 
partition, class 10 is mainly about the mathematical analogies, and class 11 is about the 
schoolboy analogies.
 
In order to find some small clusters, leading to useful relationships that are semantically 
very close due to cluster compactness, we looked at various sized partitions derived from 
the hierarchical clustering dendrogram. From a 50-cluster partition, we find the following. 

Cluster 40 consisted of the words ``mathematician'', ``poet''. 

Cluster 43 consisted of the words ``example'', ``analysis”, “algebra''. 

Cluster 47 consisted of the words ``reason'', ``mathematical''. 

Cluster 48 consisted of the words ``truths'', ``general''. 

Cluster 50 consisted of the words ``truths'', ``mathematical''. 

Cluster 15, including ``letter'' had these words: ``possession'', ``letter'', ``premises'', 
``still'', ``since'', ``observed'', ``said'', ``main'', ``far'', ``power''. 

Cluster 1 consisted of ``puff'', ``abernathy''. 

Cluster 20 consisted of the words ``document'', ``especially'', ``things'', ``point'', ``importance''.

Cluster 27 consisted of the words ``personage'', ``document'', ``royal'', ``thorough'', ``necessity'', 
``question'', ``make''. 

Our overall objectives here are to determine potentially interesting
word associations, that could then be taken as, or found to be, some
triadic metaphor (synchronic) relationship, or metonymy, a diachronic
relationship. Richardson (1985) has discussion of time dimension of
consciousness, related to diachrony. (This relates to the Vietnam War,
and is different from our work here.)

Synchrony is indicative of the contemporary presence of all three
Lacanian registers in every human act. Word-wise and textually, this
may be inferred only by means of metaphors. But consciousness may
reflect or echo the message of only one out of these three, which
therefore provides explanation in a sequential manner, parallel to the
arrow of time which characterizes human consciousness. In Iurato et
al.\ (2016), at issue is the origin of human consciousness. This is linked
to time development, that gives rise to diachrony, and we refer to this
for further examples of this type.

In very general analogy to the observational science of astronomy,
we do not seek to statistically test the properties of what is found, but
rather to obtain relevant, candidate relationships, that, as candidate
relationships, will then be assessed further in other contexts. Such, we
may wish to state, could be considered for the words ``poet'' and
``mathematician'' in this case.

\section{Conclusions}

In this work, we introduce for the first time quantitative text analysis
in humanities, and in psychoanalysis in particular, that casts a bridge
between human and what are termed natural or exact sciences

Through the semantic mapping of the storyline, we have a visualization approach for displaying 
how patterns found can be related to Lacanian registers. 

This semantic mapping is into the Euclidean metric endowed factor space. All relationships 
between the units of analysis, e.g. sentences, paragraphs, text segments, and their attributes, 
here retained word sets, i.e.\ corpora, are accounted for in the mapping into the factor space, 
so in that sense, i.e.\ taking account of all interrelationships, this is a semantic mapping. The 
endowing of the dual spaces of text units and their attributes with the Euclidean metric in the 
factor space is from the initial text unit and attribute spaces that are endowed with the chi 
squared metric. 

We have then considered approaches to the clustering of semantic and contextualized data. Beyond 
the semantics as such, the main contextualization at issue here is relating to chronology. We 
considered different chronological units, including successive sentences, successive speaker-related 
paragraphs, story segments that can be helpful for summarizing one's understanding, and for focusing 
one's interpretation. 

In the sense of unsupervised classification and exploratory data analysis, our approach is both 
``The model should follow the data, and not the reverse!'' (Benz\'ecri quotation) and ``Let the 
data speak for themselves'' (Tukey quotation). 

Our text analysis has pointed out the intertwining among three Lacanian registers. The storyline 
segments, in the semantic analysis, identify a quasi-cyclic circuit starting from the Real and 
the Imaginary registers to the Symbolic one.

In all the planar projection plots related to this semantic analysis of chronology by means of 
storyline segments, we note that Imaginary clusters are almost always placed in the centre of 
each diagram (clearly in Figure 8), besides to be the intermediate, hinge step between the Real 
(the realm of angst and fear according to Schmitz, 2015, Schmitz and Bayer, 2014) and the Symbolic 
(socio-symbolic domination of Schmitz). So the Symbolic roughly corresponds to Schmitz's 
Habitus-field intermezzo, coherently with the fact that Lacanian Symbolic corresponds to Freud's 
Super-Ego agency, the place in which there takes place the crucial passage from thing representation 
to word representation. Furthermore, we also note the prevalence of Real register in the first steps 
of semantic storyline, moving to Imaginary toward Symbolic, the prevalence of unconscious realm 
underlying conscious meaning of language. 

To aid in reproducibility of our research findings, a copy of the Poe story, as a CSV (comma 
separated value) format with sentences on successive lines, and much of the R code used in the 
analytics at issue here, have been provided at this address: http://www.narrativization.com

In conclusion, our Correspondence Analysis of Poe's story has been useful in identifying certain 
formal structures resembling the action of Lacan's registers in giving rise to language.

\section*{Acknowledgements}
To add acknowledgement for paragraph 2 of the Introduction section.

\section*{References}
\begin{enumerate}

\item 
B\'ecue-Bertaut, M., Kostov, B., Morin, A. and Naro, G. (2014). ``Rhetorical strategy in forensic 
speeches: Multidimensional statistics-based methodology'',  {\em Journal of Classification}, 31, 
85--106.

\item
Benz\'ecri, J.--P. (1973). {\em L'Analyse des Donn\'ees, Tome II Correspondances}, Paris: Dunod.

\item 
Benz\'ecri, J.--P. (1982). {\em Histoire et Pr\'ehistoire de l'Analyse des Donn\'ees}, Paris: Dunod.

\item 
Benz\'ecri, J.--P. (1983). ``L'Avenir de l'analyse des donn\'ees'', {\em Behaviormetrika}, 14, 1--11.

\item 
Blasius, J. and Greenacre, M., Eds. (2014). {\em The Visualization and Verbalization of Data}, 
Boca Raton, FL: Chapman and Hall/CRC.

\item 
Bottiroli, G. (1980). ``Strutturalismo e strategia in Jacques Lacan.
Un'interpretazione de ``La lettera rubata'' '', Aut Aut, 177--78, 95--116.

\item 
Bottiroli, G. (2006). Che cos'\`e la teoria della letteratura. Fondamenti e
problemi. Torino: Giulio Einaudi editore.

\item 
Department of English at FJU (2010). English Language Literary Criticism, Fu Jen Catholic 
University, Taiwan, ``Edgar Allan Poe, The Purloined Letter'' 
http://www.eng.fju.edu.tw/Literary\_Criticism/structuralism/purloined.html

\item 
Derrida, J. (1980). La carte postale. De Socrate \`a Freud et au-d\'el\`a.
Paris: Flammarion.

\item 
Grenfell, M. and Lebaron, F. (2014). {\em Bourdieu and Data Analysis: Methodological Principles 
and Practice}, Bern: Peter Lang. 

\item 
Husson, F., L\`e, S. and Pag\`es, J. (2011). {\em Exploratory Multivariate Analysis by Example Using 
R}, Boca Raton, FL: Chapman and Hall/CRC.

\item 
Iurato, G., Khrennikov, A. and Murtagh, F. (2016). ``Formal
foundations for the origin of human consciousness'', {\em p-Adic Numbers,
Ultrametric Analysis and Applications}, 8(4), 249--279, 2016.

\item 
Lacan, J. (1956). ``Seminar on The Purloined Letter'', 
http://www.lacan.com/purloined.htm

\item 
Lebaron, F. and Le Roux, B. (2015). {\em La M\'ethodologie de Pierre Bourdieu en Action: Espace 
Culturel, Espace Social et Analyse des Donn\'ees}, Paris: Dunod.

\item 
Le Roux, B. and Rouanet, H. (2004). {\em Geometric Data Analysis: From Correspondence Analysis 
to Structured Data Analysis}, Dordrecht: Kluwer (Springer). 

\item 
Legendre, P. and Legendre, L. (2012). {\em Numerical Ecology}, 3rd edn., Amsterdam: Elsevier.

\item 
Murtagh, F. (1985). {\em Multidimensional Clustering Algorithms}, Vienna, W\"urzburg: Physica-Verlag.

\item 
Murtagh, F. (2005). {\em Correspondence Analysis and Data Coding with Java and R}, Boca Raton, 
FL: Chapman \& Hall.

\item 
Murtagh, F., Ganz, A. and McKie, S. (2009). ``The structure of narrative: The case of film 
scripts'', {\em Pattern Recognition}, 42, 302--312.

\item 
Murtagh, F. (2012a). ``Ultrametric model of mind, I: Review'', {\em p-Adic Numbers, Ultrametric 
Analysis and Applications}, 4, 193--206.

\item 
Murtagh, F. (2012b). ``Ultrametric model of mind, II: Application to text content analysis'', 
{\em p-Adic Numbers, Ultrametric Analysis and Applications}, 4, 207--221.

\item 
Murtagh, F. (2014). ``Pattern recognition of subconscious underpinnings of cognition using 
ultrametric topological mapping of thinking and memory'', {\em International Journal of Cognitive 
Informatics and Natural Intelligence} (IJCINI), 8(4), 1--16.

\item 
Murtagh, F. and Ganz, A. (2015). ``Pattern recognition in narrative: Tracking emotional expression 
in context'', {\em Journal of Data Mining and Digital Humanities}, vol. 2015.

\item 
Poe, E.A. (1845a). ``The purloined letter'' 
http://americanliterature.com/author/edgar-allan-poe/short-story/the-purloined-letter

\item 
Poe, E.A. (1845b). {\em Tales by Edgar A. Poe}, Wiley and Putnam, New York, pp. 200--218
https://ia600408.us.archive.org/0/items/tales00poee/tales00poee.pdf

\item 
Poe, E.A. (1845c). {\em The Gift, Christmas, New Year and Birth Present}, MDCCCXLV, 
pp.\ 41--61 \\
https://ia802706.us.archive.org/2/items/giftchristmasnew00carerich/giftchristmasnew00carerich.pdf

\item
Ragland, E. (2015). {\em Jacques Lacan and the Logic of Structure. Topology and 
Language in Psychoanalysis}, New York: Routledge.

\item 
Recalcati, M. (2012--16). {\em Jacques Lacan}. 2 vols., Milano: Raffaello
Cortina Editore.

\item
Richardson, W. (1985). ``Lacanian Theory''. In: Rothstein, A. (Ed.)
(1985). {\em Models of the Mind. Their Relationships to Clinical Work}.
Madison (CT): International Universities Press, Inc., pp.\ 101--118.

\item
Ricoeur, P. (1977). {\em The Rule of Metaphor}, New York: Routledge.

\item
Schmitz, A. (2015). ``The space of Angst'', presentation (40 slides), 
{\em Empirical Investigation of Social Space II Conference}, Bonn, Germany, 
12--14 October.

\item 
Schmitz, A. and Bayer, M. (2014). ``Strukturale Psychologie: Konzeptionelle 
\"Uberlegungen und empirische Analysen zum Verh\"altnis von Habitus und Psyche'' 
(``Structural psychology: Conceptual considerations and empirical analyses on the 
relationship of habitus and psyche''), preprint, 17 pp.

\item 
Todd, J.M. (1990). {\em Autobiographics in Freud and Derrida}. London:
Routledge.

\end{enumerate}

\end{document}